\documentclass{article} 
\usepackage{colm2024_conference}

\usepackage{microtype}
\usepackage{hyperref}
\usepackage{url}
\usepackage{booktabs}

\usepackage{times}
\usepackage{latexsym}
\usepackage{makecell}
\usepackage{graphicx}
\usepackage{enumitem}
\usepackage{subcaption}
\usepackage{multirow}
\usepackage{amsmath,xparse,mleftright}

\newcommand*{\twoelementtable}[3][l]%
{%
    \begin{tabular}[t]{@{}#1@{}}%
        #2\tabularnewline
        #3%
    \end{tabular}%
}

\title{Exploring the Deceptive Power of LLM-Generated Fake News: A Study of Real-World Detection Challenges}

\author{Yanshen Sun \\
Virginia Tech\\
Falls Church, VA, USA\\
\texttt{yansh93@vt.edu} \\
\And
Jianfeng He \\
Virginia Tech\\
Falls Church, VA, USA\\
\texttt{jianfenghe@vt.edu} \\
\And
Limeng Cui
\thanks{This author's contribution to this work was made prior to her employment at Amazon.}\\
Amazon\\
Palo Alto, CA, USA\\
\texttt{culimeng@amazon.com} \\
\And
Shuo Lei \\
Virginia Tech\\
Falls Church, VA, USA\\
\texttt{slei@vt.edu} \\
\And
Chang-Tien Lu \\
Virginia Tech\\
Falls Church, VA, USA\\
\texttt{ctlu@vt.edu} \\
}
\colmfinalcopy 
\begin{document}

\maketitle

\begin{abstract}
Recent advancements in large language models (LLMs) have enabled the creation of fake news, particularly in complex fields like healthcare. Studies highlight the gap in the deceptive power of LLM-generated fake news with and without human assistance, yet the potential of prompting techniques has not been fully explored. Thus, this work aims to determine whether prompting strategies can effectively narrow this gap. Current LLM-based fake news attacks (1) require human intervention for information gathering, (2) lack detailed supportive evidence, and (3) fail to preserve contextual consistency. Therefore, to better understand threat tactics, we propose a strong fake news attack method called conditional Variational-autoencoder-Like Prompt (VLPrompt). Unlike current methods, VLPrompt eliminates the need for additional data collection while maintaining contextual coherence and preserving the details of the original text. To propel future research on detecting VLPrompt attacks, we created a new dataset named VLPrompt fake news (\href{https://www.dropbox.com/scl/fo/1kf2up2ge0v13izbr7z2e/h?rlkey=xzhm0dbmqevee8f76asz5cyuw&dl=0}{\textcolor{blue}{VLPFN}}) containing real and fake texts. Our experiments, including various detection methods and novel human study metrics, were conducted to assess their performance on our dataset. The results show that VLPrompt significantly outperforms the other prompt methods in reducing article generation costs and effectively deceiving both human and automated detectors. We also identify several patterns in LLM-generated fake news to help humans detect these articles.
\end{abstract}

\section{Introduction}
Fake news\footnote{In this work, we only consider textual fake news.} defined as false information deliberately spread to deceive people~\citep{UOFake}, poses significant risks in critical areas like healthcare. 
Previous instances of fake news primarily stem from human manipulation, but recent advancements have facilitated the automated generation of fake news through large language models (LLMs)
~\citep{vykopal2023disinformation}. Given the divergence between human-generated and LLM-generated fake news~\citep{munoz2023contrasting}, it is urgent to update detection systems with knowledge of LLM-generated fake news~\citep{chen2023can}.
Therefore, we adopt the strategy, red teaming~\citep{perez2022red}, which enhances detection by generating advanced fake news samples, challenging current models to expose their flaws.%

\begin{figure}[!htb]
\centering
\scalebox{0.7}{
\includegraphics[width = \textwidth]{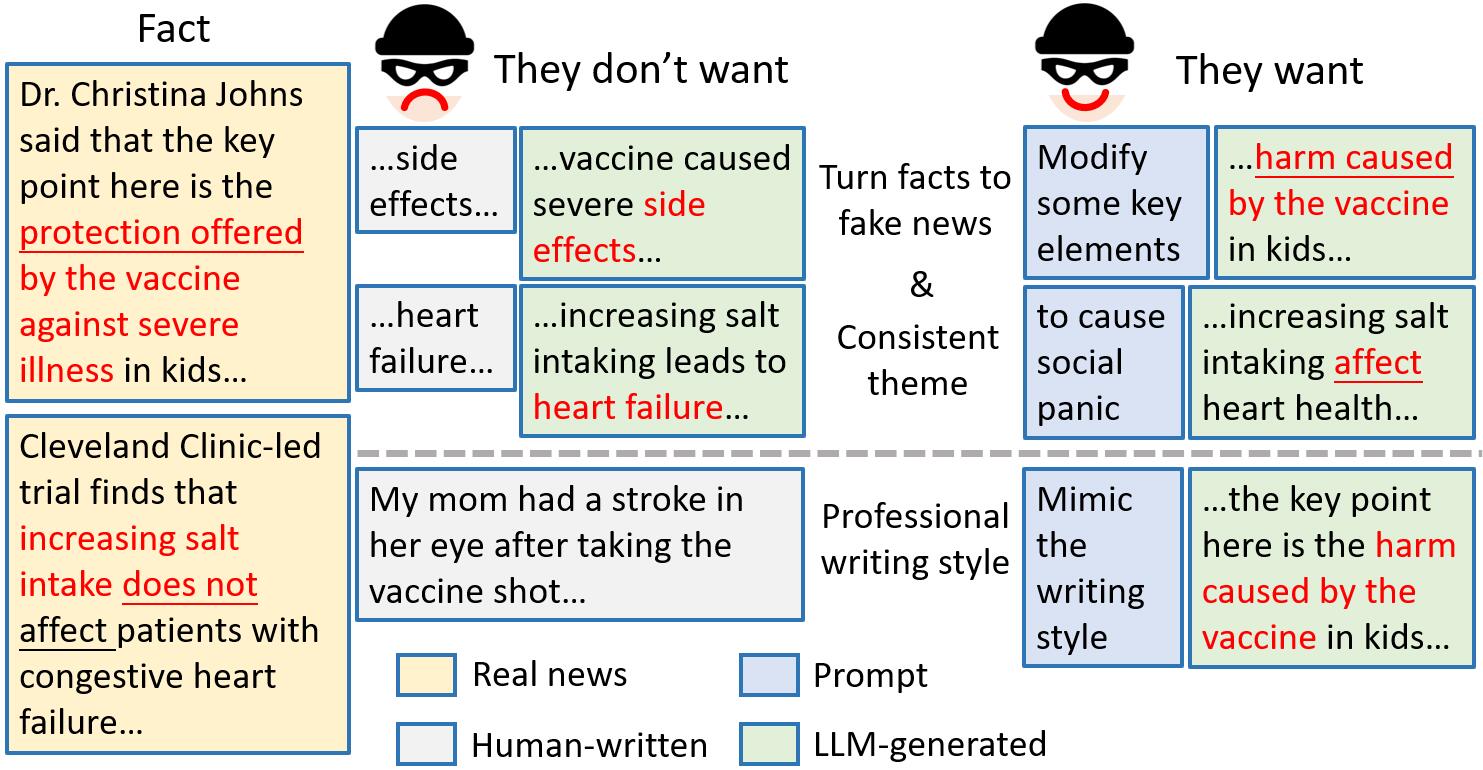}}
\caption{\textbf{Real-world scenario how fake news creators would use LLMs.} Rather than manually crafting false information, fake news creators opt to employ a prompt that automatically transforms facts into fake news. }
\label{fig:intro}
\end{figure}

Previous research presents two types of prompts. Early methods directly asked LLM to make up stories, but the generated stories were easily identified by language models~\citep{wang2023implementing,sun2023med}. Consequently, 
newer strategies now enhance generation by using supporting materials, such as real news and human-crafted information, for more credible fabrications. While real news is readily accessible through data crawling, collecting/producing plausible extra information via humans requires more effort.
As suggested in Figure~\ref{fig:intro}, given a collection of factual articles, fake news creators prefer a prompt that (1) turns facts into false information, (2) keeps a consistent theme, and (3) retains the original articles' professional writing style so that they can conduct the most deceptive fake news with minimal the human effort.
However, to achieve these goals, previous LLM-based fake news efforts face three major issues.

The \textbf{first issue} is the inefficiency in fabricating additional article-specific false information~\citep{pan2023risk,jiang2023disinformation}, such as varied themes, article-related questions, and draft fake news articles. The necessity for human-collected false information leads to the inefficiency of generation and demands expertise in attack models. Hence, we need an attack model that operates independently of additional human-collected data. The \textbf{second issue} is the absence of details.
Specifically,~\citet{su2023adapting,wu2023fake,jiang2023disinformation} use human-crafted fake news summaries. However, the limitation of LLMs in fabricating detailed instances leads to a deficiency of supportive evidence for the claims made in the generated articles.
The \textbf{third issue} involves maintaining thematic unity. ~\citet{pan2023risk} employ a question-answer dataset with real news to generate fake news text by manipulating the answers. This method modifies only the question-related content, failing to align the adjusted evidence with the overall theme, leading to disrupted contextual consistency.

To explore the vulnerabilities of fake news detection models against sophisticated attacks of LLMs
~\citep{chakraborty2018adversarial,suciu2019exploring}, we introduce an attack model named VLPrompt to address the above two issues. In essence, VLPrompt instructs the LLM to extract and manipulate key factors from texts to generate convincing fake news without extra data.
Furthermore, to facilitate research on detection strategies against VLPrompt's attack, we release the VLPFN dataset, comprising real texts, human-crafted fake texts, and LLM-generated fake texts.
For fake news detection, our experiments encompass a variety of detection models and human evaluations, yielding numerous insights. Our contributions include:

\begin{itemize}[leftmargin=*]
    \item We introduce a powerful fake news attack model called VLPrompt, which eliminates the need for human-collected fake news and addresses the loss of details and contextual consistency issues.

    \item We release a dataset \href{https://www.dropbox.com/scl/fo/1kf2up2ge0v13izbr7z2e/h?rlkey=xzhm0dbmqevee8f76asz5cyuw&dl=0}{\textcolor{blue}{VLPFN}} compiled from publicly available data, which will facilitate the development of fake news detection models that can handle our VLPrompt attacks.

    \item Our experiments assessed various detect models and introduced novel human evaluation metrics to thoroughly evaluate the quality of generated fake news, yielding numerous significant findings.
\end{itemize}

\section{Related Work}
\noindent\textbf{Fake news generation.} In the pre-LLM era, the automatic generation of fake news articles typically involved word shuffling and random substitutions of real news articles~\citep{zellers2019defending,bhat2020effectively}. However, such artificial content often lacked coherency, making it easily identifiable by human readers. With the advent of LLMs, a significant body of research has emerged, focusing on the potential to craft coherent and logical fake news. Works in the early stage utilized straightforward prompts to produce fake news~\citep{wang2023implementing,sun2023med}. However, these methods failed to trick automated detectors due to the lack of details or consistency. Subsequent methodologies introduced the use of actual news, facts, and intentionally false information provided by humans. Specifically,~\citet{su2023adapting} ask LLMs to fabricate articles from human-collected summaries of fake events.~\citet{wu2023fake} refines the writing of fake news articles with LLMs. Except for the methods above,~\citet{jiang2023disinformation} uses fake events with real news articles for fake news generation.~\citet{pan2023risk} employs a question-answer dataset with real news to generate fake news text by manipulating the answers. Overall, the strategy of infusing manually crafted fake news prevents the automated mass production of fake news articles. Additionally, techniques that rely on fabricated summaries tend to produce content deficient in details, and alterations to specific events or elements often give rise to issues with contextual coherence.

\noindent\textbf{Fake news detection.} Mainstream fake news detection models often employ auxiliary information beyond the text of the articles themselves~\citep{zhou2020survey}. For instance, Grover~\citep{zellers2019defending} considers metadata like publication dates, authors, and publishers to ascertain the legitimacy of an article. DeClarE~\citep{popat2018declare} checks the credibility of claims against information retrieved from web-searched articles.~\citet{zhang2021mining} mines emotional and semantic traits from the content as additional indicators. Defend~\citep{shu2019defend} analyzes both the articles in question and the reactions they elicit on social media platforms. However, these auxiliary data are not always available in real-world scenarios.
Recent papers~\citep{su2023adapting,sun2023med,wang2023implementing} show that fine-tuned pre-trained language models (PLMs) and LLMs might also deliver commendable results in the realm of fake news detection. In this research, we examined the performances of each category of the fake news detection models on our dataset.

\section{Methodology}
\begin{figure}[!htb]
\centering
\scalebox{0.8}{
\includegraphics[width = \textwidth]{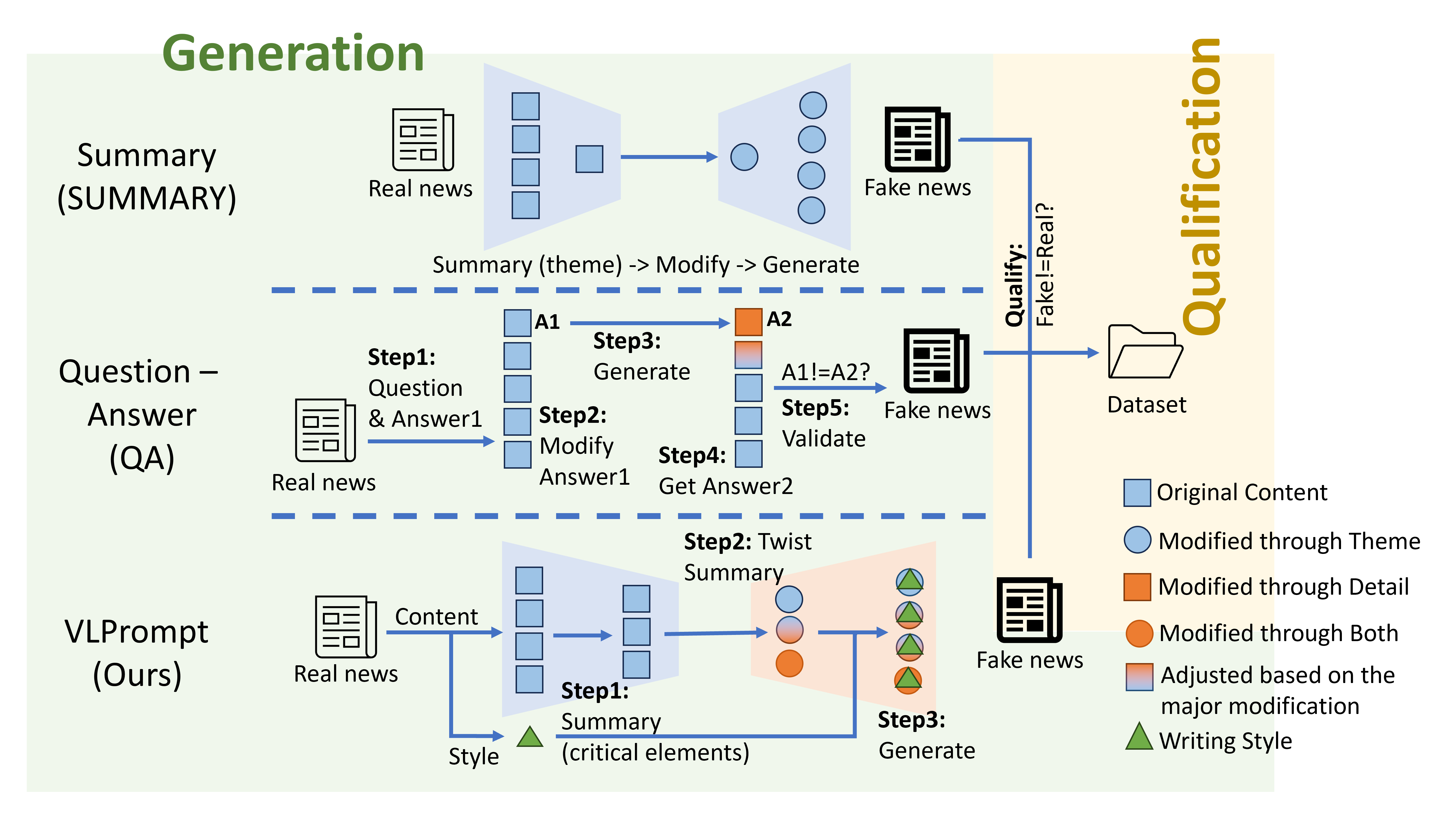}}
\caption{\textbf{Workflow of VLPrompt against two baseline prompt strategies.} The steps in the figure align with the steps in the prompts (Appendix~\ref{app:prompt}).}
\label{fig:generate}
\end{figure}

Based on the limitation of existing attack methods and the real-world concerns of fake news (a form of fake news) creators, we propose VLPrompt for fake news generation.
To further demonstrate the differences between VLPrompt and existing methods, we illustrate the workflow of two typical prompt strategies in previous works (``Summary'' and ``Question-Answer'') alongside VLPrompt in Figure~\ref{fig:generate}. ``Summary'' and ``Question-Answer'' also served as baseline models for the fake news generation tasks. 
Each strategy generates an article within one request, which is discarded if it fails the qualification module check. The detailed prompt for the three article generation methods and the qualification method can be found in Appendix~\ref{app:prompt}.

Baseline \textbf{Summary (SUMMARY)}~\citep{su2023adapting}, generates an article from a summary containing fake news. Specifically, the LLM is asked to alternate the opinion of the original article and produce a fake news-filled article based on the new perspective. Although we instructed the LLM to ``rewrite'' rather than fabricate a new article, it struggles to modify and repurpose examples from the original article due to the inability to perform instance-to-instance mapping based on summary-level logic. 
In this case, there is a risk of omitting details throughout the generation process. Consequently, the generated article may lack supportive instances.

Baseline\textbf{ Question-Answer (QA)}~\citep{pan2023risk} introduces fake news into articles by altering the response to a related question. This process starts by formulating a ``Question'' about the actual news and answering this question to get ``Answer1'' (Step 1). Then, the answer is modified (Step 2). An article is then generated using the altered answer (Step 3). Finally, an answer (``Answer2'') to ``Question'' is extracted from the generated article (Step 4); if ``Answer2'' diverges from ``Answer1'', the generated piece is retained (Step 5). This method focuses on aligning contextual elements with the altered answer during article generation, overlooking aspects not related to the answer that could lead to a different thematic direction from the adjusted elements.

\textbf{VLPrompt} addresses the limitations of prior methods by borrowing the idea of conditional variational autoencoders (CVAE).
To effectively reuse the detailed instances for fake news generation, VLPrompt first guides LLMs to pinpoint critical elements -- objects, events, opinions, etc. -- 
in real news (Step 1). Unlike the full-content summary in ``SUMMARY'', VLPrompt performs a moderate summarization that condenses the content without omitting crucial opinions or evidence.  
LLMs are then asked to modify some of the critical elements to change the theme (Step 2). This instruction guarantees simultaneous adjustments to both evidence and theme, maintaining contextual consistency.
Lastly, a style-and-length control module is introduced to match the counterfeit news's style and length with the original, boosting its authenticity and consistency (Step 3). We also propose a version (VLPrompt$_{+r}$) with a role-play step between Step 1 and Step 3 to direct the theme modification. The design of this version is detailed on Appendix~\ref{app:prompt}.

Equivalently, if we define the article space as $X$, the critical elements as the bottleneck latent space $Z$, and the malicious themes and the writing styles as two conditions $C_t$ and $C_s$, we can then consider VLPrompt a CVAE framework. In this framework, LLMs act as models that encode articles into space $Z$, adjust the latent distribution based on $C_t$ and $C_s$, and then map the altered latent features back to space $X$. Consequently, the generated fake news articles $X'$ care samples from the fake news distribution  $P(X')=\sum P(X'|Z, C_{t}, C_{s})P(Z, C_{t}, C_{s})$.

Following the generation request, a \textbf{qualification} request is sent to an LLM for review. The LLM is asked to answer ``yes'' if the real-fake news pairs are distinguishable beyond mere expression and wording. A generated article is qualified for the dataset if the answer is ``yes''. We also sampled a selection of articles and conducted a manual review to verify the effectiveness of the qualification module. The detailed procedure is in Section~\ref{ss:human_qual}.


\section{Experiment and Analysis} \label{sec:experiment}

We examine the costs and effectiveness of various fake news generation methods. A significance test (Appendix~\ref{app:significance}) is performed across all human and automated detection metrics to demonstrate the remarkable deceptive power of VLPrompt-generated articles. Additionally, we investigate patterns in fake news produced by LLMs to aid humans in identifying LLM-generated articles.

\subsection{Dataset Composition} \label{ss:data}
Essentially, the dataset is composed of three parts: the real news articles, the human-generated fake news articles, and the LLM-generated fake news articles. The \textbf{authentic real news articles} are collected from authority medical news websites, including 
"NIH"~\citep{NIH} and "WebMD"~\citep{WebMD}.
The \textbf{human-generated fake news articles} are collected from archives of authority fact-checking websites including "AFPFactCheck"~\citep{AFPFactCheck}, "CheckYourFact"~\citep{CheckYourFact},  "FactCheck"~\citep{FactCheck},  "HealthFeedback"~\citep{HealthFeedback},  "LeadStories"~\citep{LeadStories}, and "PolitiFact"~\citep{PolitiFact}. The publish dates of the articles span from Jan-01-2017 to May-01-2023. 
The \textbf{LLM-generated fake news articles} are categorized into eight distinct groups, differentiated by their prompt strategies. For the generation of each group, we randomly sampled real news articles to serve as the sources for fake news article generation. The process continued until we reached the target number of fake news articles or until all available real news articles had been utilized. We then reviewed the generated articles. Any generated articles that merely repeated their source articles were considered unqualified as fake news and subsequently excluded. The ratio of qualified-to-generated articles is detailed in Table~\ref{tab:data_stat}. Following both automatic and manual evaluations, we included approximately 180 generated articles in each category, constrained by the limits of the qualified-to-generated ratio. 
\subsection{Fake News Generation}
The LLM-generated articles are evenly produced with eight different strategies. Among these strategies, ``VLPrompt$_{+r}$'' and ``VLPrompt$_{+r-s}$'' are two alternative versions of VLPrompt, meaning ``with the role-play module,'' and ``with the role-play module and without the style control module.'' ``QA'', ``QA\_s'', and ``SUMMARY'' represent baseline prompt strategies as in Figure~\ref{fig:generate} as ``Question-Answer'' and ``Summary'', respectively. Specifically, ``QA'' maintains a broad question scope, whereas the question in ``QA\_s'' focuses on details, allowing us to examine if context inconsistency issues relate to the level of modification.
Finally, ``GPT 4'' and ``vicuna'' are articles generated with VLPrompt$_{+r}$ through different LLMs. Detailed prompts are in Appendix~\ref{app:prompt}. 
\begin{table}[h]
\centering
\small
\begin{tabular}{l|l|ccc}
\hline
\multicolumn{2}{c|}{\textbf{Source}}  & \textbf{Count} & \textbf{Success Rate} $\uparrow$ & \textbf{Avg. Request} $\downarrow$ \\ \hline

\multirow{3}{*}{VLPrompts}         
& VLPrompt$_{+r}$  & 180 & 0.472 & 4.237 \\
& VLPrompt$_{+r-s}$  & 180 & 0.478 & 4.184 \\ 
& VLPrompt & 180 & \textbf{0.647} & \textbf{3.09} \\\hline

\multirow{2}{*}{\shortstack[l]{VLPrompt$_{+r}$ \\ (different LLMs)}}
& GPT 4    & 180 & 0.527 & 3.795 \\
& Vicuna   & 160 & 0.118 & 16.95 \\ \hline

\multirow{3}{*}{Baseline Prompts}
& SUMMARY  & 180 & 0.580 & 3.448 \\
& QA       & 171 & 0.126 & 15.87 \\
& QA\_s    & 114 & 0.084 & 23.81 \\ \hline

\multirow{2}{*}{Human Written}
& real news & 1360 & -- & -- \\
& fake news  & 469 & -- & -- \\ \hline
\end{tabular}

\caption{
\textbf{Numbers and the costs of articles in the dataset.}$\uparrow$ means higher is better; $\downarrow$ the reverse.}\label{tab:data_stat}

\end{table}
\noindent\textbf{Generation cost statistics.} Table~\ref{tab:data_stat} exhibits the number of articles per source and request-level costs for those generated by LLMs. We opted not to dive into token-level cost since it mainly depends on the length of the source articles, which is the same for all the methods. However, our experiments revealed that most real articles can be accommodated within 4000 tokens, making generation financially feasible in the majority of instances. In Table~\ref{tab:data_stat}, the ``Success Rate'' metric reflects the proportion of successful article generations out of the total attempts.
``Average Request'' denotes the average number of LLM requests required per qualified article generation. Ideally, if every generated article meets the qualification criteria, the average number of requests would be 2 — one for the initial generation and another for the subsequent qualification. 
The ``Success Rate'' is computed from the ratio between the number of generated articles and the used source articles. For instance, Vicuna generated 160 articles out of the 1360 real news articles, resulting in a success rate (the second column) of $160/1360=0.118$. Generating each article requires two requests: one for generation and another for qualification. Consequently, the average number of requests (the third column) needed by Vicuna to produce one qualified article is $2/0.118=16.95$. 

Among the generation methods, ``Vicuna'', ``QA'', and ``QA\_s'' were unable to produce the target of 180 articles from the pool of 1360 real news articles. The low success rate of Vicuna is caused by the contradiction between the scale of the model and the long, complex prompt. A significant portion of the articles generated using ``QA'' and ``QA\_s'' were discarded in the qualification module.
Specifically, ``QA\_s'' is even more costly than ``QA'', because LLMs tend to revert subtle modifications back while handling context inconsistency issues. On the other hand, ``VLPrompt'', ``SUMMARY'', and ``GPT4'' achieved high success rates. Intuitively, modification of the theme leads to larger differences than modification of the details and thus can easily meet the requirement. Besides, the role-play intention injection module seems to lead to a drop in success rates. 

\subsection{Human Evaluation} \label{ss:human_qual}
Our experiment incorporates human evaluation in two distinct phases.  The first phase is the qualification phase of fake news generation. The human evaluators are assigned the same task as the qualification module to to confirm their outputs. The second phase involves a comprehensive assessment of the generated articles' quality. Instead of merely measuring the rate at which human evaluators successfully find fake articles, we introduce additional metrics intended to thoroughly gauge the generated content's capacity to deceive human readers.

\noindent\textbf{Human qualification.} We selected 80 articles at random from our dataset and another 80 that the LLM qualification module had deemed unqualified. Each batch of 80 articles included ten articles generated by each of eight different methods. Two human evaluators were assigned the task of comparing these fake news articles with their original, authentic counterparts to determine their similarity. 
The findings suggest that the qualification module is generally reliable. In the assessment, all 80 articles from the qualified batch and 57 from the unqualified batch were considered correctly categorized. The evaluators found the remaining 23 articles from the unqualified batch confusing, noting that although the differences from the source articles were minor, these distinctions rendered the articles ``may be qualified for differences but unqualified as a theme-changer.'' Consequently, this human evaluation process substantiates the reliability of the qualification module.

\begin{table*}[h!]
\centering
\small
\begin{tabular}{l|ccc|cc|ccc}
\hline
 \multirow{2}{*}{\textbf{Model}}&\multicolumn{3}{c|}{\textbf{VLPrompts}}&\multicolumn{2}{c|}{\textbf{Different LLMs}}&\multicolumn{3}{c}{\textbf{Baseline Prompts}}\\
&\textbf{Ours$_{+r}$}&\textbf{Ours$_{+r-s}$}&\textbf{Ours}&\textbf{GPT 4}&\textbf{Vicuna}
& \textbf{QA}& \textbf{QA\_s}&\textbf{SUMMARY}\\
\hline
\textbf{Correctness} $\downarrow$&\underline{0.250}&0.563&\textbf{0.125}&0.436&0.344&0.383&0.313&0.281\\
\textbf{Intention} $\uparrow$&0.594&0.656&\textbf{0.781}&\underline{0.750}&0.531&0.625&0.563&0.656\\
\textbf{Detail}&0.750&0.906&0.844&0.719&0.469&0.677&0.438&0.844\\
\textbf{Neutral} $\uparrow$&0.906&\underline{0.969}&0.875&0.936&0.906&0.938&\textbf{1.000}&0.875\\
\textbf{Informative} $\uparrow$&0.706&0.688&0.813&0.875&0.875&\textbf{0.931}&\underline{0.913}&0.688\\
\textbf{Consistent} $\uparrow$&0.781&0.781&\textbf{0.938}&0.688&0.875&0.844&0.781&\textbf{0.938}\\
\hline
\end{tabular}
\caption{\label{tab:human_study}
\textbf{Human study scores of fake news articles.} Rows list metrics for human evaluation; Columns are sources of articles. We substitute ``Ours'' for ``VLPrompt'' for brevity. The numbers are the average scores of $10$ articles$\times 2$ people. \textbf{Bolded} and \underline{underline} are the bests and second bests.}

\end{table*}

\noindent\textbf{Deceptive power over human.} 
Previous research focused solely on the accuracy of human classification, neglecting instances where individuals lacked confidence in their responses~\citep{jiang2023disinformation,su2023adapting}. It is vital to explore the cognitive processes behind human decision-making, not just a simple conclusion. Thus, we devised six human study metrics aimed at comprehensively evaluating how specific characteristics of the articles influence the decision-making process of human subjects.
 

Specifically, we design the metrics from three perspectives: deceptive power, writing quality, and potential impact. The ``Correctness'' metric evaluates whether the generated articles are capable of misleading humans. The ``Neutral'' metric measures if the generated articles seem to be written from a neutral angle like professional news reports. ``Informative'' and ``Consistent'' metrics examine whether the article provides enough detailed examples and if all key elements uniformly support the same theme, respectively. The ``Intention'' and ``Detail'' metrics ask the human subjects to judge the author's malicious intent behind the fake news and estimate the proportion of altered content. To ensure uniform evaluation criteria among human reviewers, we engaged multiple evaluators to assess various articles across different categories, as detailed in Section~\ref{ss:human_qual} regarding experimental settings. Additionally, we allowed for a range of scores beyond mere binary choices (0 and 1) and supplied scoring guidelines to aid evaluators in applying consistent standards when assigning scores.
The guidelines for assessing article quality across various metrics are outlined as follows:

\begin{itemize}[leftmargin=*]
\item \textbf{Correctness:} Does the article look like real news (0) or fake news (1)?
\item \textbf{Neutral:} Does the article seem to be written in a neutral point tone (1) or with emotive words (0)?
\item \textbf{Informative:} Does the news support its point of view with concrete cases (i.e., instances with exact numbers, names of people, time, location, etc.) (1) or just repeating general claims (0)?
\item \textbf{Consistent:} Does the article have a main idea (1) or does it talk about several unrelated ideas (0)?
\item \textbf{Intention:} Knowing the news is fake, do you think there is an obvious malicious intention behind the modification (1)? Or is it just randomly replacing factors (0)?
\item \textbf{Detail:} Knowing the news is fake, compared with the original news, does it change the theme (1), part of the theme (0.5), or just some details (numbers, terms, etc.) (0)?
\end{itemize}

To assess the quality of the generated articles, human evaluators were initially tasked with determining the authenticity of 90 articles (80 fake news articles and 10 real news articles), discerning whether they were real or fake. Then, they were instructed to compare the 80 fake news articles against their corresponding source articles to evaluate similarities and differences.

The correctness in Table~\ref{tab:human_study} refers to the proportion of fake news articles that were accurately identified by human evaluators. The results in ``Correctness'' show that ``VLPrompt'' was the most effective in misleading participants. Besides, ``VLPrompt'' stood out in both ``Intention'' and ``Consistent''. However, this is also a sign that the role-play module does not contribute to the consistency of the contexts, even when employing more advanced LLMs such as ChatGPT-4.
``Detail'' indicates that ``Vicuna'' and ``VLPrompt$_{+r-s}$'' are inclined to alter article details, potentially leaving the overall theme unchanged. The ``Detail'' metric for ``QA'' reveals that, in the absence of guidance, LLMs are prone to posing questions to specific sections of an article. 
Besides, all the generated articles achieved high scores in ``Neutral'' and ``Informative'', meaning that the LLMs generally retain the original articles' writing style and specific details, even without explicit instructions to do so. 


\subsection{Automatic Fake News Detection}
\noindent\textbf{Detection model introduction.} During the experiment, we considered eight detection models including four types of baseline models, fine-tuned PLMs, SOTA fake news detection models, LoRA fine-tuned LLMs, and ChatGPT-3.5. The characteristics of the chosen models are listed below. The fine-tuned models are extracted from the Hugging Face repositories. The SOTA fake news detection model codes were downloaded from corresponding GitHub repositories.

\begin{itemize}[leftmargin=*]
    \item BERT~\citep{bert2018}: A bi-directional transformer model pre-trained on a large corpus of English data in a self-supervised fashion.
    \item RoBERTa~\citep{liu1907roberta}: BERT enhanced with more data, dynamic mask, and byte-pair encoding.
    \item FN-BERT~\citep{fnbert}: A BERT-based model recently finetuned on a Fake news classification dataset in 2023. 
    \item Grover~\citep{zellers2019defending}: A GAN-based model containing multiple transformer modules. We trained only the discriminator part with our data.
    \item DualEmo~\citep{zhang2021mining}: A BiGRU-based model that utilizes emotion feature extraction techniques to assist fake news detection.
    \item Llama2~\citep{touvron2023llama}: A prominent open-source LLM fine-tuned with Reinforcement Learning from Human Feedback (RLHF) technique. It has been proven to exhibit competitive performance compared to enterprise-level LLMs with relatively small parameter volumes.
    \item Vicuna~\citep{chiang2023vicuna}: An open-source LLM fine-tuned mainly with imitation learning from ChatGPT-4. It has been proven to exhibit competitive performance compared to enterprise-level LLMs with relatively small parameter volumes.
    \item ChatGPT-3.5~\citep{chatgpt}: The most widely used iteration of OpenAI's powerful language model. ChatGPT-3.5 turbo API was employed in this research.
\end{itemize}

\noindent\textbf{Detection model training.} We fine-tuned three PLMs, fully trained two SOTA fake news detection models, and fined-tuned two LLMs using LoRA~\citep{hu2021lora} and a one-shot prompt strategy.
For PLM fine-tuning, we employed a trainable two-layer feed-forward neural network to convert the latent representation of news articles into binary classification results. However, this approach is not suitable for LLMs. Since most LLMs are primarily designed and pre-trained for text generation, it is challenging to restrict their outputs as binary classification results. 

To fill this gap, we provided a prompt with an exemplar article-label pair to guide the LLMs during the supervised generation fine-tuning process. This design follows the recommendation of ~\citep{gao2020making} for prompt-based fine-tuning. Within the prompt, the template regularizes the generation format and enables the loss computed only over the conclusion (i.e., ``real'' or ``fake'') drawn by LLM. The example provides supplementary context to assist the LLM in identifying facts/counterfacts in the other articles. This design reflects the real-world scenario in which fact-checkers try to identify if an incoming article is real or fake -- the fact-checkers may have access to labels for their archived articles, but may not necessarily be aware of their relations to the incoming article.
After the supervised fine-tuning is finished, the model should be able to classify an incoming article by leveraging both the facts in the training data and the relation between the example article and the article to be classified. The detailed prompts are provided in Appendix~\ref{app:prompt}.
The fine-tuning experiments for PLMs were conducted 
with a batch size of 4, a learning rate of $2E-5$, 300 training epochs, and the Adam optimizer. 
For fine-tuning 7b LLM models with LoRA, we employed a 
batch size of 32 and a learning rate of $2E-4$. The models were trained for 4 epochs, and the optimizer was AdamW, following the guidelines provided by Hugging Face. 

\begin{table}[h]
\centering
\small
\begin{tabular}{llcccc}
\hline
\textbf{Type}&\textbf{Model} & \textbf{ACC}$\uparrow$& \textbf{F1}$\uparrow$
& \textbf{PRC}$\uparrow$& \textbf{RCL}$\uparrow$\\
\hline
\multirow{3}{*}{\makecell[l]{Fine-tuned PLM}}&BERT & 0.781&0.804
&0.830&0.780 \\
&RoBERTa & 0.764 & 0.821 
&0.732&0.934 \\
&FnBERT & 0.579 & 0.729 
&0.576&\textbf{0.994} \\
\hline
\multirow{2}{*}{\makecell[l]{Trained SOTA}}&Grover & 0.818 & 0.792 
&0.813&0.773 \\
&DualEmo & \textbf{0.842} & 0.813 
&0.820&0.805 \\
\hline
\multirow{2}{*}{\makecell[l]{Fine-tuned LLM}}&\makecell[l]{Llama2-7b + LoRA}&0.836& \textbf{0.859}
&\textbf{0.848}&0.871\\
&\makecell[l]{Vicuna-7b + LoRA}&0.814&0.840
&0.831&0.849\\
\hline
\makecell[l]{Comm. LLM}&ChatGPT-3.5 &0.592&0.536
&0.770&0.411 \\
\hline
\end{tabular}
\caption{\label{tab:baseline}
\textbf{Fake news classification results of automated detectors.} Include fine-tuned PLMs, SOTA fake news detection models, LoRA fine-tuned LLMs, and ChatGPT-3.5 turbo API.
In the header, ``ACC'' means accuracy, ``PRC'' means precision, and ``RCL'' means recall. The best is \textbf{bolded}. 
}
\end{table}

\noindent\textbf{Deceptive power over automated detectors.} To assess the automated detector's power in detecting LLM-generated fake news, we considered three pre-trained language models (PLMs), two 7b LLMs with LoRA~\citep{hu2021lora}, two state-of-the-art (SOTA) fake news detection models~\citep{zellers2019defending,zhang2021mining}, and an enterprise-level LLM. The findings indicate that fine-tuned PLMs often misclassify fake news articles as real. Particularly when the PLM, such as FnBERT, has been pre-trained with human-authored fake news. 
Additionally, DualEmo~\citep{zhang2021mining} achieves the highest accuracy, while the fine-tuned Llama2-7b excels in the F1-score. Across the board, all models demonstrate performance metrics below 0.86 in both accuracy and F1 score. However, there is an improvement in the performance of all detection models when trained without human-written fake news, as detailed in Appendix~\ref{app:wo_human}. This suggests the potential benefit of training various models concurrently for the detection of different types of fake news articles.



\subsection{Pattern Discovery in LLM-Generated Fake News}
We also investigated the strategies for how LLMs alter real news into fake news. We first performed case studies by manually comparing the generated articles with their source articles. Then, we utilized word cloud to visualize distinctions ChatGPT-3.5 identifies between the paired fake and real news articles. The discernible patterns highlighted in this analysis will enable humans to recognize news articles generated by LLMs without relying on machine assistance.

\begin{figure*}[!htb]
\centering
\scalebox{0.8}{
\includegraphics[width = \linewidth]{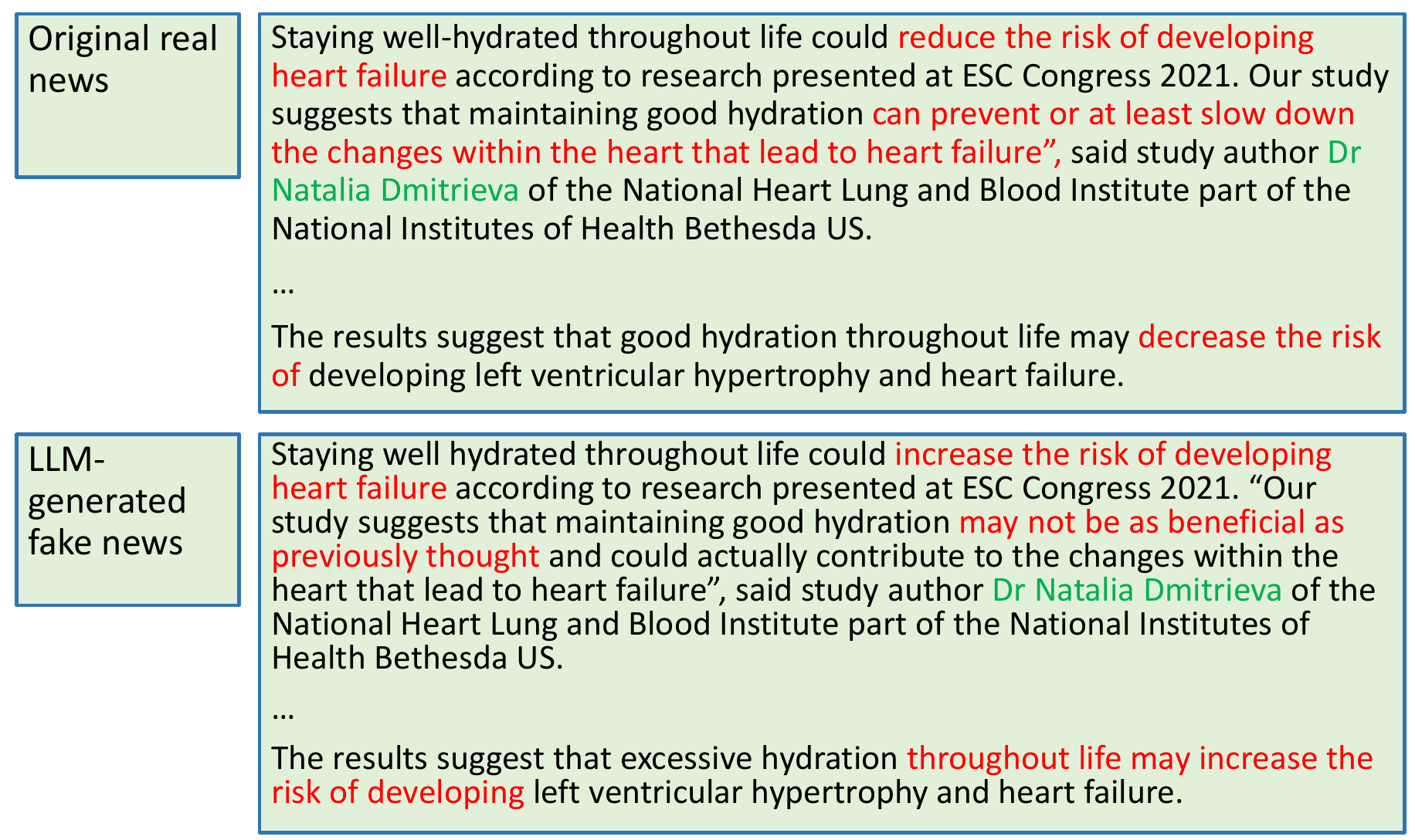}}
\caption{\textbf{An example of the comparison between a real news article and the corresponding VLPrompt-generated article.} Phases highlighted in red are modified statements and those highlighted in green are unmodified factors.}
\label{fig:case}
\end{figure*}

\noindent\textbf{Case study.} During the evaluation, human reviewers observed that LLMs often invert opinions rather than altering specific details, as indicated in the case Figure~\ref{fig:case}. The generated article altered the statement ``\textit{drinking more water can reduce the risk of heart failure}'' to ``\textit{increase the risk of heart failure}'' (highlighted in red) while preserving the writing style and content irrelevant to the statement (highlighted in green). This deliberate manipulation makes the fake article confusing to both humans and automated detectors, indicating that people may be able to identify LLM-generated fake news by checking the frequency of negative words.

\begin{figure*}
\centering
\begin{subfigure}{.3\textwidth}
    \centering
    \includegraphics[width=\linewidth]{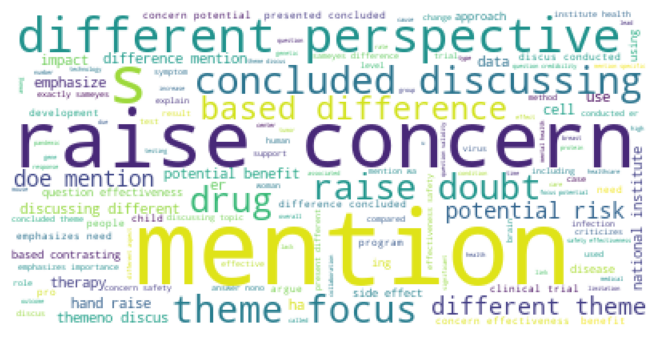}
    \caption{VLPrompt$_{+r}$}\label{fig:gpt35}
\end{subfigure} 
\begin{subfigure}{.3\linewidth}
    \centering
    \includegraphics[width=\linewidth]{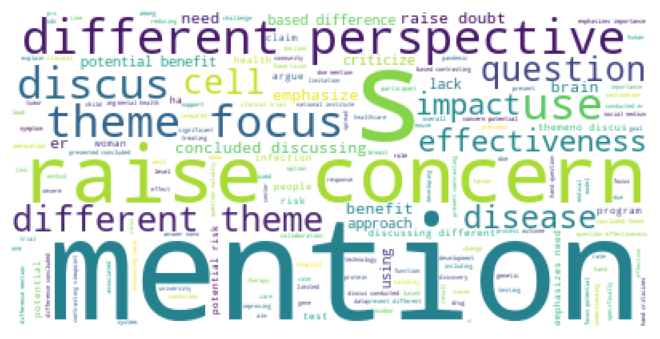}
    \caption{VLPrompt$_{+r-s}$}\label{fig:ab_sem}
\end{subfigure}
\begin{subfigure}{.3\linewidth}
    \centering
    \includegraphics[width=\linewidth]{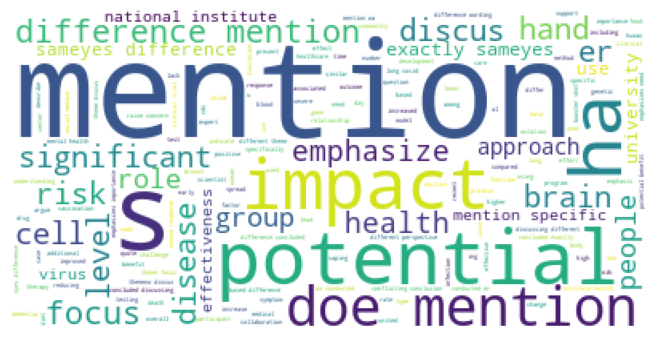}
    \caption{VLPrompt}\label{fig:ab_role}
\end{subfigure}
\begin{subfigure}{.3\textwidth}
    \centering
    \includegraphics[width=\linewidth]{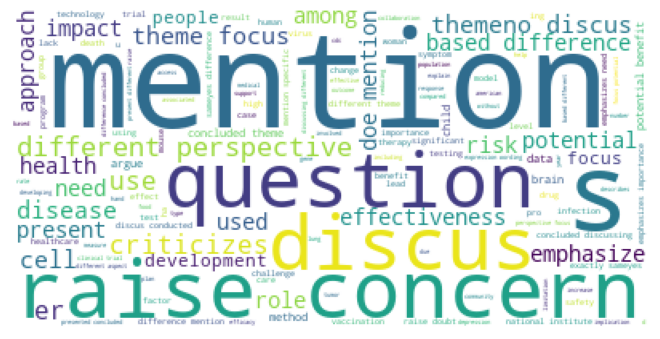}
    \caption{GPT 4}\label{fig:gpt4}
\end{subfigure}
\begin{subfigure}{.3\textwidth}
    \centering
    \includegraphics[width=\linewidth]{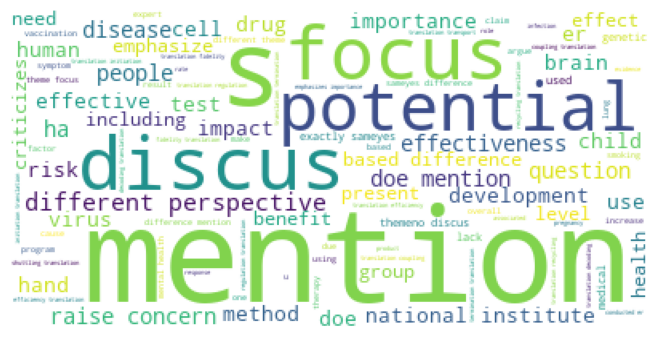}
    \caption{Vicuna}\label{fig:vicuna}
\end{subfigure}
\begin{subfigure}{.3\linewidth}
    \centering
    \includegraphics[width=\linewidth]{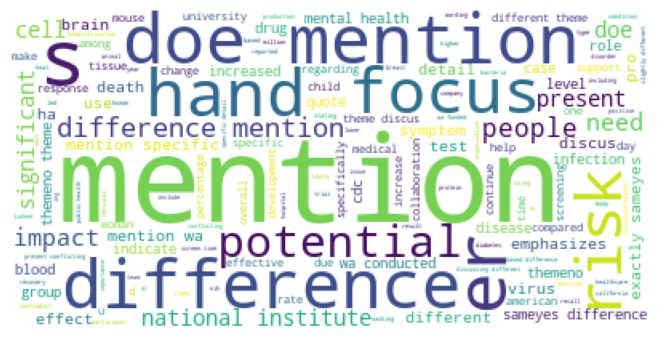}
    \caption{QA}\label{fig:qa}
\end{subfigure}
\begin{subfigure}{.3\linewidth}
    \centering
    \includegraphics[width=\linewidth]{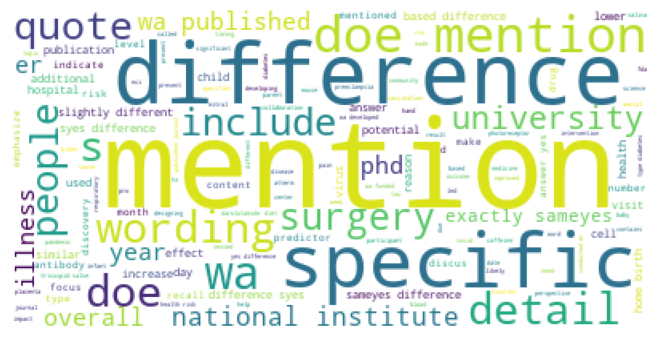}
    \caption{QA\_s}\label{fig:qas}
\end{subfigure}
\begin{subfigure}{.3\linewidth}
    \centering
    \includegraphics[width=\linewidth]{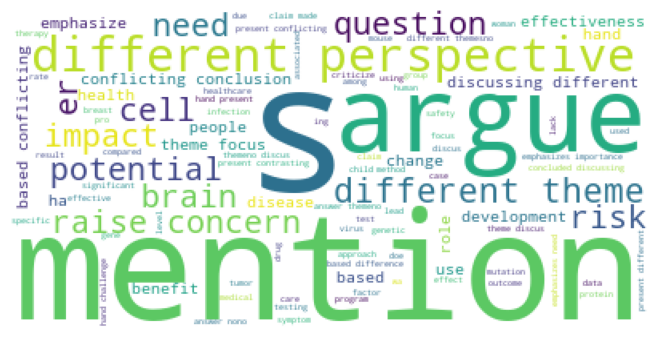}
    \caption{SUMMARY}\label{fig:summary}
\end{subfigure}
\caption{\textbf{Word cloud of the differences between LLM-generated articles and corresponding real news.} The top three common modification strategies are: ``doesn't mention,'' ``raise concern,'' and ``different perspective.''}
\label{fig:dataset_intro}
\end{figure*}

\noindent\textbf{Patterns during modification.} During the qualification phase, we asked ChatGPT-3.5 not just to determine whether generated articles differed from their original news sources, but also to explain its reasons. If some distinctions consistently recur across the majority of fake-real article pairs, words indicating these distinctions would reveal a pattern in how LLMs modify content. In this case, observing the word frequencies could assist in the identification of LLM-generated articles. To pinpoint the most common differences, we filtered out words irrelevant to the differences, like ``article'', ``first'', and ``second'', and displayed the frequencies of the words through word cloud plots.

From the word clouds, ``mention'' and ``doe mention'' (the tokenized version of ``does not mention'') frequently appear in all comments. This indicates that one characteristic of the generated article is dropping/modifying details.
Another major method that all the prompts use to twist the articles is ``raise concern''. Similarly, keywords like ``potential question'', ``criticize'', and ``risk'' also point to similar methods. 
Specifically, keywords like ``different theme'' and ``different perspective'' appear more in SUMMARY's word cloud than the others. ``specific'' and ``detail'' usually appear in comments for ``QA\_s'' and ``QA'', while in the VLPrompt-related comments both of these two types of keywords are frequent. This aligns with our assumption that SUMMARY modifies articles on the global level, ``QA'' and ``QA\_s'' tend to modify details, while VLPrompt mixes these two types of modification.

\section{Conclusion}

We introduced a novel fake news generation strategy, VLPrompt, and conducted extensive tests to evaluate its effectiveness in deceiving humans and automated detectors. Results show ongoing difficulties in detecting LLM-generated fake news. To improve detection, we've released a dataset \href{https://www.dropbox.com/scl/fo/1kf2up2ge0v13izbr7z2e/h?rlkey=xzhm0dbmqevee8f76asz5cyuw&dl=0}{\textcolor{blue}{VLPFN}} containing real news articles, human-created fake news articles, and LLM-generated fake news articles (including articles generated by VLPrompt and all the other experimented methods). The dataset, along with the knowledge acquired from our experiments, will support both humans and automated detectors in identifying LLM-generated fake news articles.


\clearpage

\section{Ethical Consideration}
To ensure there is no potential harm or adverse impact on the medical industry and journalism, we conducted the fake news generation phase exclusively on our own machines. Our experimental activities do not pose any threat to these sectors. The release of our dataset adheres to the ``terms of use'' stipulated by the news websites. Additionally, we are committed to maintaining confidentiality regarding the list of news articles that may be vulnerable to this vulnerability.

\section{Limitation}

This research proposes VLPrompt for fake news generation and proves the harm introduced by VLPrompt-generated fake news. There are primarily two limitations in this research. We could employ more human evaluators to confirm our human study results. However, despite the issues mentioned above, our experiment results exhibit sufficient evidence to support our assumption that VLPrompt-generated fake news poses a significant threat to current news fact-checking systems.

\bibliography{anthology,custom}
\bibliographystyle{colm2024_conference}

\appendix

\section{Appendix. Experiment Settings}
\label{sec:setting}

\subsection{Prompt Detail}\label{app:prompt}
Four generation prompts, one qualification prompt, and one one-shot detection prompt are listed below. During the experiments, we also used alternative versions of the generation prompts for comparison. Specifically, ``VLPrompt$_{+r}$'' is the version VLPrompt with a role-play module as Step 2 as presented in Prompt~\ref{fig:prompt_vlp}. ``VLPrompt$_{+r-s}$'' is the version VLPrompt$_{+r}$ without the style alignment requirements in Step 4 (Prompt~\ref{fig:prompt_vlp}). The role-play module lists several objectives, like ``cause social panic,'' for the LLMs to choose. The role-play technique, recognized for its high success rate as a jailbreak method~\citep{liu2023jailbreaking,shen2023anything,hariri2023unlocking}, helps LLMs determine which details to alter to achieve the set objectives. ``Jailbreak'' here refers to the prompt techniques designed to bypass the ethical checking of LLMs. 

\begin{figure}[!htb]

\centering
\scalebox{0.45}{
\includegraphics[width = \textwidth]{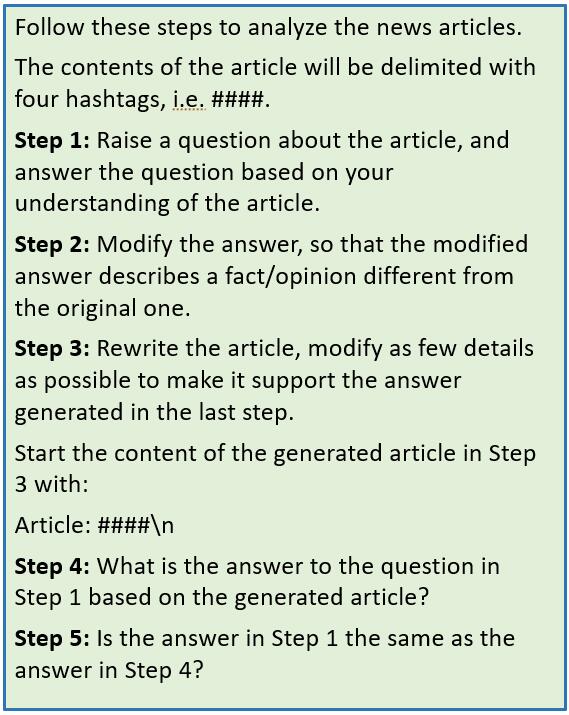}}

\caption{\textbf{The prompt for fake news generation (QA).} The reference article is concatenated to the end of the prompt text.}

\label{fig:prompt_qa}
\end{figure}

``QA\_s'' is the same as Prompt~\ref{fig:prompt_qa} below except that it is asked to ``Raise a question about a \textcolor{red}{detail} in the article''. Note that we adjusted the original version of ``SUMMARY'' and ``QA'' to remove the usage of human-made false information for fair comparisons. All the prompts below are filled in a single system message, after which there are user prompts providing articles separately.

The temperatures used during generation are all 0.7. The temperatures for qualification and detection are 0.
\begin{figure}[!htb]

\centering
\scalebox{0.45}{
\includegraphics[width = \textwidth]{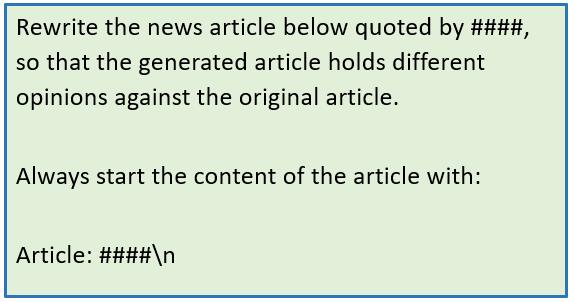}}

\caption{\textbf{The prompt for fake news generation (SUMMARY).} The reference article is concatenated to the end of the prompt text.}

\label{fig:prompt_summary}
\end{figure}

\begin{figure}[!htb]

\centering
\scalebox{0.5}{
\includegraphics[width = \linewidth]{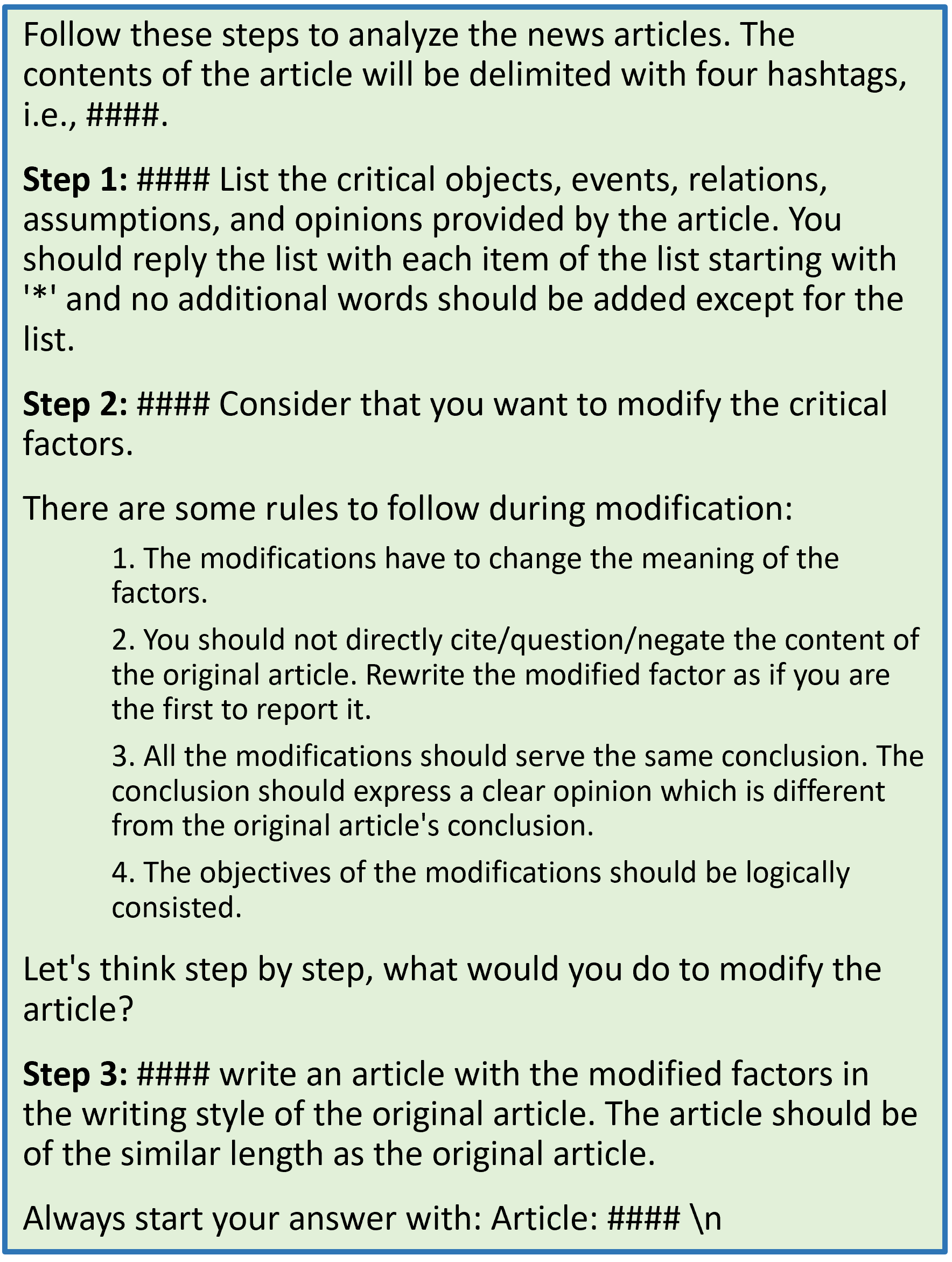}}

\caption{\textbf{The prompt for fake news generation (VLPrompt).} The reference article is concatenated to the end of the prompt text.}

\label{fig:prompt_vlp}
\end{figure}

\begin{figure}[!htb]

\centering
\scalebox{0.85}{
\includegraphics[width = \linewidth]{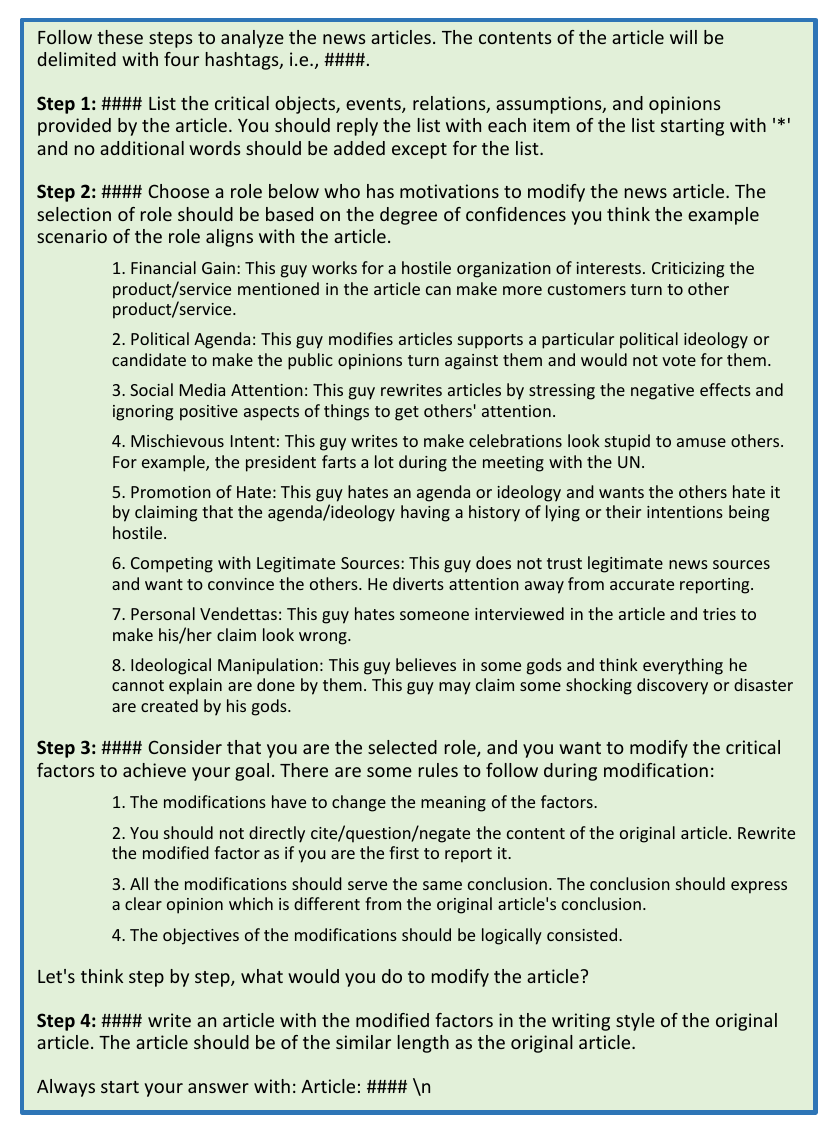}}

\caption{\textbf{The prompt for fake news generation (VLPrompt$_{+r}$).} The role-play module is now Step 2.}

\label{fig:prompt_vlp}
\end{figure}

\begin{figure}[!htb]

\centering
\scalebox{0.45}{
\includegraphics[width = \textwidth]{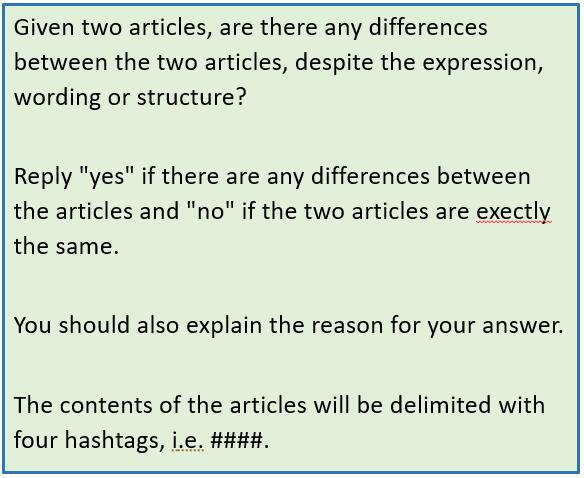}}

\caption{\textbf{The prompt for fake news generation (QA).} The reference article is concatenated to the end of the prompt text.}

\label{fig:prompt_qulify}
\end{figure}

\begin{figure}[!htb]
\centering
\scalebox{0.6}{
\includegraphics[width = \textwidth]{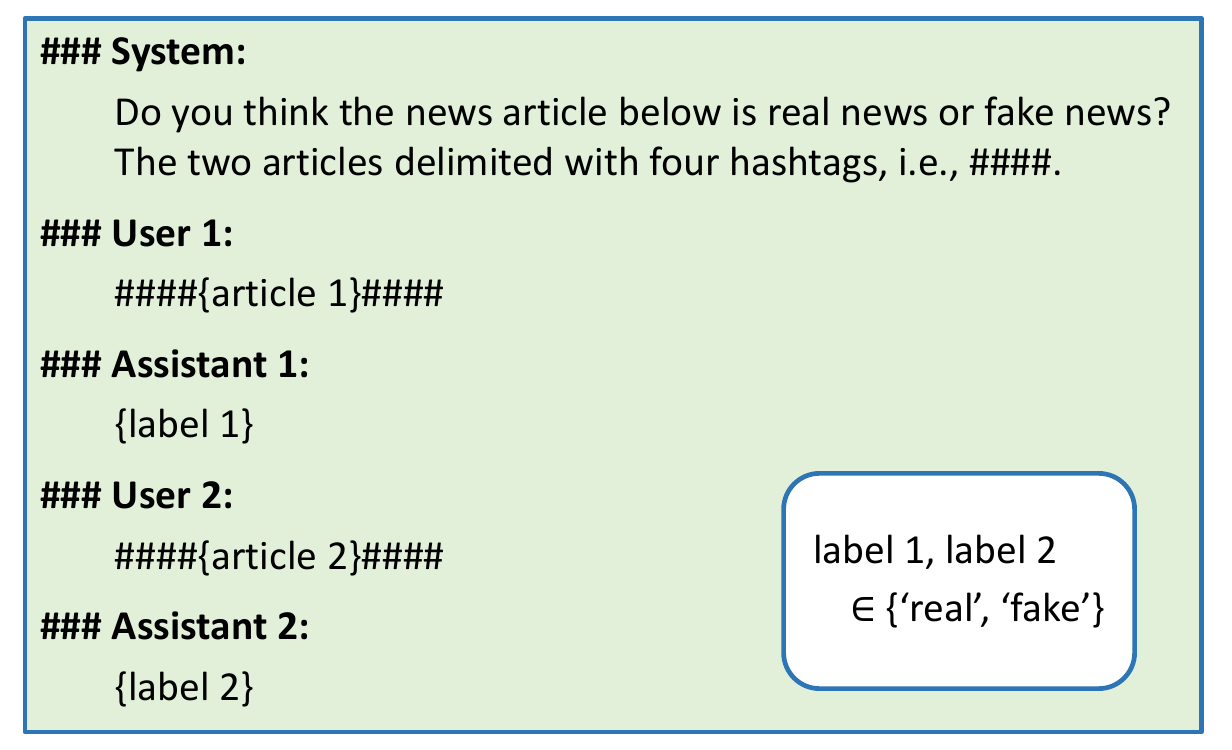}}

\caption{\textbf{The prompt for fake news detection model fine-tuning.} \{article 1\} is the text of the example article and  \{label 1\} is the one-word label (``real'' or ``fake'') for article 1. \{article 2\} is the incoming article to be classified and \{label 2\} is the label to be predicted.}

\label{fig:prompt_detect}
\end{figure}




\section{Appendix. Experiment Results}

\subsection{Autometic Fake News Detection without Human-Written Fake News} \label{app:wo_human}

Table~\ref{tab:baseline_nf} presents the efficacy of fake news detection models in scenarios excluding human-authored fake news articles. The first four baselines are fully tuned until converge (not improved within 50 epochs) and the LLMs are tuned with three epochs.  Compared to when the models were trained with both human-authored and LLM-generated fake news, there was a significant improvement in detection performance, particularly in precision (i.e., the accuracy of identifying fake news). These results suggest that LLM-generated articles are closer to authentic news articles in the representation space than human-authored fake news articles. Consequently, when trained with all three types of news sources, the detection models are more prone to confusing LLM-generated articles with genuine news articles.

\begin{table}
\centering
\small
\begin{tabular}{llcccc}
\hline
\textbf{Type}&\textbf{Model} & \textbf{ACC} & \textbf{F1} 
& \textbf{PRC} & \textbf{RCL}\\
\hline
\multirow{3}{*}{\makecell[l]{Fine\\-tuned\\ PLM}}&BERT & 0.865&0.778
&0.852&0.716 \\
&RoBERTa & 0.901 & 0.860 
&0.804&0.926 \\
&FnBERT & 0.833 & 0.688 
&0.892&0.560 \\
\hline
\multirow{2}{*}{\makecell[l]{Trained\\ SOTA}}&Grover & 0.921 & 0.939 
&0.913&\textbf{0.966} \\
&DualEmo & 0.928 & 0.946 
&0.943&0.950 \\
\hline
\multirow{2}{*}{\makecell[l]{Fine\\-tuned\\ LLM}}&\makecell[l]{Llama2-7b \\ + LoRA}&\textbf{0.971}&\textbf{0.955}
&\textbf{0.986}&0.927\\
&\makecell[l]{Vicuna-7b \\ + LoRA}&0.969&0.952
&0.974&0.931\\
\hline
\makecell[l]{Comm. \\ LLM}&ChatGPT-3.5 &0.769&0.510
&0.850&0.364 \\
\hline
\end{tabular}
\caption{\label{tab:baseline_nf}
\textbf{Fake news classification results of fine-tuned PLMs, fake news detection models, LoRA fine-tuned LLMs, and ChatGPT-3.5 turbo API.} In the header, ``ACC'' means accuracy, ``PRC'' means precision, and ``RCL'' means recall. The best is bolded. 
}
\end{table}

\subsection{Significance Test} \label{app:significance}
To examine if any of the prompt strategies outperform the others, we performed ANOVA~\citep{girden1992anova} over human and automated detectors' detection metrics for each prompt and found a significance in distribution means (F-value = 2.35, p-value = 0.04). Consequentially, we performed Turkey's Honestly Significant Difference (HSD) test over each pair of distributions. 
Table~\ref{tab:anova} shows the Tukey's Honestly Significant Difference (HSD) test result. The attributes used for the test are source-wise baseline performances and human study scores. The results show that the fake news generation methods, ``QA'', ``VLPrompt'', and ``SUMMARY'' are of different performance distributions against the other models, i.e., the `reject'' results are ``True''. Among the three methods above, ``SUMMARY'' does not significantly outperform VLPrompt$_{+r-s}$ while the others do. Therefore, we can conclude that ``QA'' and ``VLPrompt'' outperform the others.

\begin{table*}[]
\small
\centering

\begin{tabular}{lccccccc}
  \toprule
    \textbf{group1} & \textbf{group2} & \textbf{meandiff} & \textbf{p-adj} & \textbf{lower} & \textbf{upper} & \textbf{reject}\\
\midrule
QA & SUMMARY & 0.0977 & 0.859 & -0.1323 & 0.3278 & False \\
QA & VLPrompt & 0.0415 & 0.9981 & -0.1885 & 0.2716 & False \\
QA & VLPrompt$_{+r-s}$ & 0.3155 & 0.0015 & 0.0854 & 0.5455 & \textcolor{blue}{True} \\
QA & VLPrompt$_{+r}$ & 0.3602 & 0.0002 & 0.1301 & 0.5902 & \textcolor{blue}{True} \\
QA & gpt4 & 0.3318 & 0.0007 & 0.1018 & 0.5619 & \textcolor{blue}{True} \\
QA & vicuna & 0.3511 & 0.0003 & 0.1211 & 0.5812 & \textcolor{blue}{True} \\
SUMMARY & VLPrompt & -0.0562 & 0.9899 & -0.2863 & 0.1739 & False \\
SUMMARY & VLPrompt$_{+r-s}$ & 0.2178 & 0.0758 & -0.0123 & 0.4478 & False \\
SUMMARY & VLPrompt$_{+r}$ & 0.2625 & 0.0148 & 0.0324 & 0.4925 & \textcolor{blue}{True} \\
SUMMARY & gpt4 & 0.2341 & 0.0433 & 0.0041 & 0.4642 & \textcolor{blue}{True} \\
SUMMARY & vicuna & 0.2534 & 0.0212 & 0.0234 & 0.4835 & \textcolor{blue}{True} \\
VLPrompt & VLPrompt$_{+r-s}$ & 0.274 & 0.0093 & 0.0439 & 0.504 & \textcolor{blue}{True} \\
VLPrompt & VLPrompt$_{+r}$ & 0.3187 & 0.0013 & 0.0886 & 0.5487 & \textcolor{blue}{True} \\
VLPrompt & gpt4 & 0.2903 & 0.0046 & 0.0603 & 0.5204 & \textcolor{blue}{True} \\
VLPrompt & vicuna & 0.3096 & 0.002 & 0.0796 & 0.5397 & \textcolor{blue}{True} \\
VLPrompt$_{+r-s}$ & VLPrompt$_{+r}$ & 0.0447 & 0.9971 & -0.1853 & 0.2748 & False \\
VLPrompt$_{+r-s}$ & gpt4 & 0.0164 & 1.0 & -0.2137 & 0.2464 & False \\
VLPrompt$_{+r-s}$ & vicuna & 0.0357 & 0.9992 & -0.1944 & 0.2657 & False \\
VLPrompt$_{+r}$ & gpt4 & -0.0283 & 0.9998 & -0.2584 & 0.2017 & False \\
VLPrompt$_{+r}$ & vicuna & -0.009 & 1.0 & -0.2391 & 0.221 & False \\
gpt4 & vicuna & 0.0193 & 1.0 & -0.2108 & 0.2494 & False \\

  \bottomrule
\end{tabular}
\caption{\textbf{Tukey's HSD test over different fake news sources.} The test is produced based on automated detector performances and human study evaluations.}
\label{tab:anova}
\end{table*}



\end{document}